\pdfoutput=1
\documentclass[11pt]{article}

\usepackage[]{ACL}

\usepackage{times}
\usepackage{latexsym}

\usepackage[T1]{fontenc}
\usepackage[utf8]{inputenc}

\usepackage{microtype}

\usepackage{inconsolata}

\usepackage{user}

\title{BAND: Biomedical Alert News Dataset}

\author{Zihao Fu$^{\spadesuit}$\ \ \ \  \textbf{Meiru Zhang}$^{\spadesuit}$\ \ \ \   \textbf{Zaiqiao Meng}$^{\spadesuit \diamondsuit}$\ \ \ \  \\
 \textbf{Yannan Shen}$^{\heartsuit}$\ \ \ \ 
 \textbf{David Buckeridge}$^{\heartsuit}$\ \ \ \ 
\textbf{Nigel Collier}$^\spadesuit$ \\
$^\spadesuit$Language Technology Lab, University of Cambridge \\
$^\diamondsuit$School of Computing Science, University of Glasgow \\
$^{\heartsuit}$School of Population and Global Health, McGill University\\
 \texttt{$^\spadesuit$\{zf268, mz468, nhc30\}@cam.ac.uk}, $^{\diamondsuit}$zaiqiao.meng@glasgow.ac.uk\\
 $^{\heartsuit}$\{yannan.shen david.buckeridge\}@mail.mcgill.ca}

\begin{document}
\maketitle
\begin{abstract}
    Infectious disease outbreaks continue to pose a significant threat to human health and well-being. To improve disease surveillance and understanding of disease spread, several surveillance systems have been developed to monitor daily news alerts and social media. However, existing systems lack thorough epidemiological analysis in relation to corresponding alerts or news, largely due to the scarcity of well-annotated reports data. To address this gap, we introduce the Biomedical Alert News Dataset (BAND)\footnote{\url{https://github.com/fuzihaofzh/BAND}}, which includes 1,508 samples from existing reported news articles, open emails, and alerts, as well as 30 epidemiology-related questions. These questions necessitate the model's expert reasoning abilities, thereby offering valuable insights into the outbreak of the disease. The BAND dataset brings new challenges to the NLP world, requiring better disguise capability of the content and the ability to infer important information. We provide several benchmark tasks, including Named Entity Recognition (NER), Question Answering (QA), and Event Extraction (EE), to show how existing models are capable of handling these tasks in the epidemiology domain. To the best of our knowledge, the BAND corpus is the largest corpus of well-annotated biomedical outbreak alert news with elaborately designed questions, making it a valuable resource for epidemiologists and NLP researchers alike.
\end{abstract}

\section{Introduction}
In spite of advancements in healthcare, infectious disease outbreaks continue to pose a substantial threat to human health and well-being. To enhance disease surveillance and deepen our understanding of disease transmission, several surveillance systems have been established, including BioCaster \citep{meng2022biocaster}, GPHIN \citep{mawudeku2013global}, ProMED-mail \citep{yu2004promed}, and HealthMap \citep{freifeld2008healthmap}, EIOS\footnote{\url{https://www.who.int/initiatives/eios}}. These systems monitor daily news alerts and social media platforms, providing real-time surveillance and analysis of disease outbreaks.

\begin{figure}[t]
    \centering
    \includegraphics[width=0.5\textwidth]{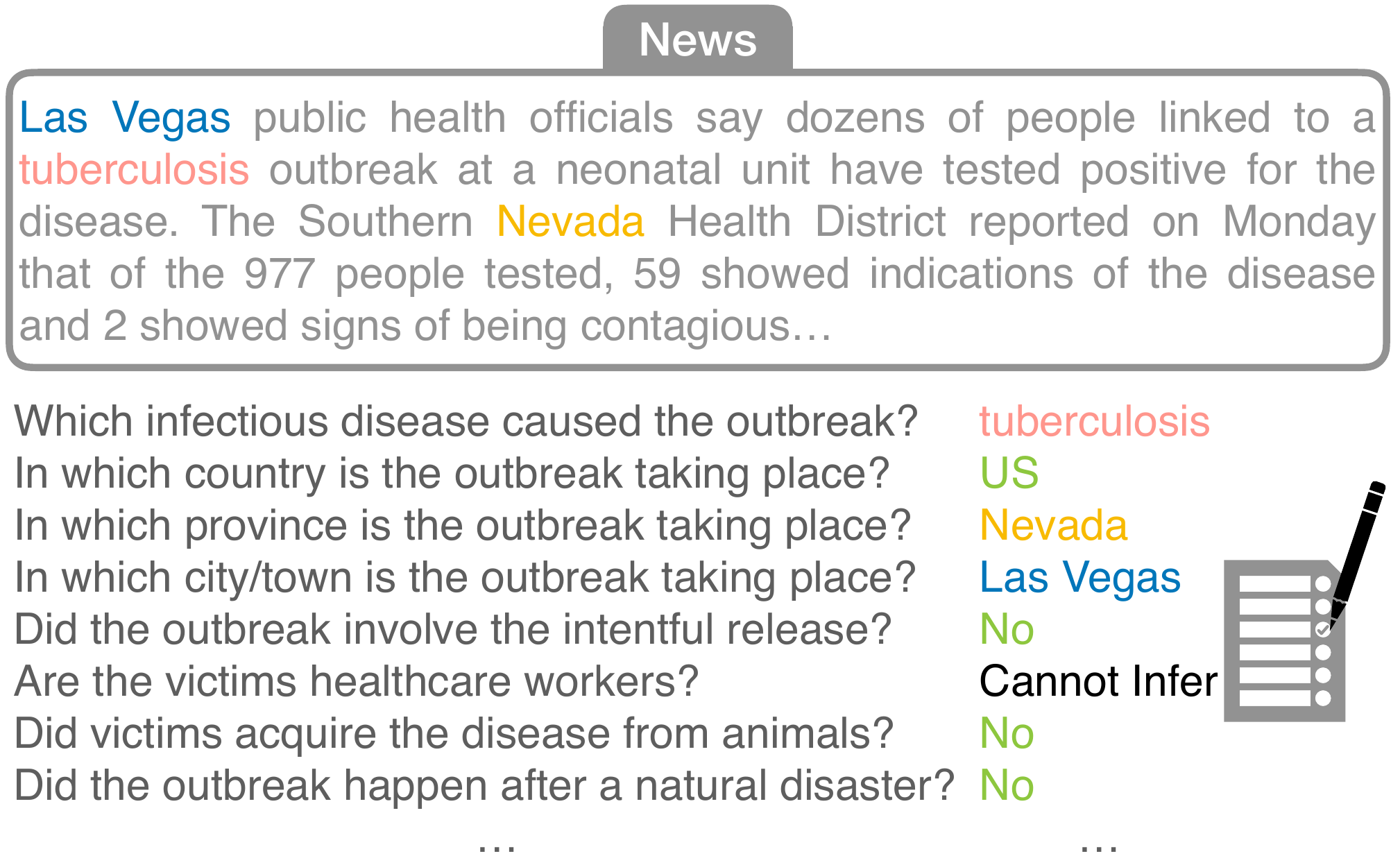}
    \caption{The BAND dataset is annotated by answering 30 specially designed by epidemiology experts. Annotators are required to infer answers (highlighted in \textcolor{mygreen}{green}) based on their background knowledge when the information is not explicitly provided in the given text.}
    \label{fig:sample}
\end{figure}

Despite the notable achievements of current surveillance systems, most of them concentrate on the detection of outbreak events using social media and news sources. However, there is a dearth of systems that offer automatic and comprehensive epidemiological analysis of corresponding alerts or news. For instance, epidemiologists require such systems that can identify cases where the disease is deliberately released or affects vulnerable populations, such as the elderly or children. The automatic identification of these cases can facilitate preventive measures and prompt rescue efforts. The limited capacity of existing systems can partly be attributed to the scarcity of well-annotated report data, which is critical for training machine learning systems for domain experts. Although several existing datasets \citep{torres2022global, carlson2023world} have been annotated to extract outbreak events, they mostly focus on the statistics (e.g. location, disease names, and etc.) of the outbreak event rather than providing a thorough epidemiological analysis for further investigation.

To enrich the capabilities of existing surveillance systems, we present a newly annotated dataset, namely the Biomedical Alert News Dataset (BAND). This dataset comprises 1,508 samples extracted from recently reported news articles, open emails, and alerts, accompanied by 30 epidemiology-related questions. These questions cover most of event-related queries raised in \citet{torres2022global, carlson2023world} as well as more detailed inquiries regarding the outbreak event. For instance, we annotate whether an outbreak was an intentional release or involved a pregnant woman (refer to \Cref{tab:questions} for details of all the questions), which are important risk factors considered by human public health analysis. Affirmative responses to these questions serve as indicators for epidemiologists to prioritize and assess the need for further action. The selection of samples and questions are meticulously carried out by domain experts specializing in the fields of epidemiology and NLP. This dataset aims to empower NLP systems to analyze and address several critical questions, which aids the current surveillance systems in identifying significant trends and providing insights on how to improve disease surveillance and management.

This dataset presents new challenges for the NLP community, particularly in the area of common sense reasoning. For example, as illustrated in \Cref{fig:sample}, the system must automatically extract the outbreak country from the given news, even when it is not explicitly stated. This requires the model to infer the country name from context clues such as city name, state name, and report organization. In addition, the dataset requires better content disambiguation capabilities. For instance, when asked to identify the city of the outbreak, many cities worldwide share the same name, making it necessary to provide a geocode\footnote{\url{https://www.geonames.org/}} to uniquely identify the location. 
Our datasets can be used to assess the capabilities of state-of-the-art models across a range of benchmark NLP tasks.
In particular, we have performed experiments on three prominent tasks, including Question Answering (QA), Named Entity Recognition (NER), and Event Extraction (EE), to showcase the effectiveness of current models in addressing these tasks on this new dataset.

The contribution of this paper can be summarized as follows:

1) We introduce the BAND corpus, which is the largest corpus of well-annotated biomedical alert news with elaborately designed questions to the best of our knowledge.

2) We provide various model benchmarks for a range of NLP tasks, including Named Entity Recognition (NER), Question Answering (QA), and Event Extraction (EE).

3) We present a complete pipeline for annotating biomedical news data that can be leveraged for annotating similar datasets.

\section{The BAND Corpus}
\begin{figure}[t]
    \centering
    \includegraphics[width=0.5\textwidth]{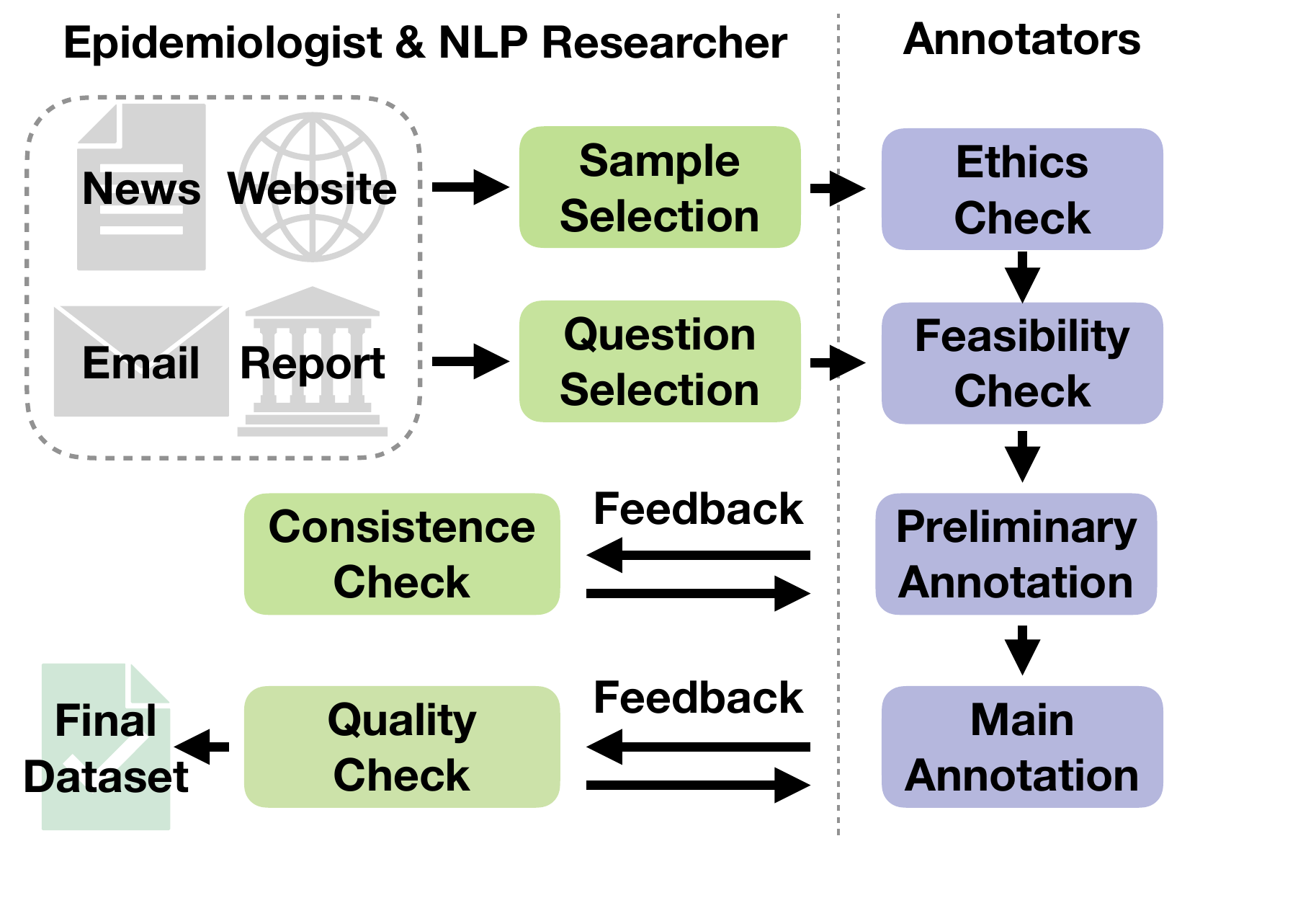}
    \caption{Our annotation workflow.}
    \label{fig:workflow}
\end{figure}

The BAND corpus consists of 1,508 authentic biomedical alert news articles and 30 expert-generated questions aiming at enhancing understanding of disease outbreak events and identifying significant incidents requiring special attention. The alert news articles encompass a wide range of sources, including publicly available news articles, emails, and reports. The annotation process is outlined in  \Cref{fig:workflow}. Initially, epidemiology and NLP researchers select appropriate questions and samples from the alert news. Experienced annotators then conduct an ethics check to filter out unsuitable content and perform a feasibility check to ensure that they can annotate the selected questions. Should any questions arise, annotators consult with the experts for clarification. Subsequently, annotators engage in a preliminary annotation process by answering the selected questions for each sample. Following this, a consistency check is performed to ensure the same understanding among all annotators regarding the questions and samples. Finally, in the main annotation stage, annotators proceed to annotate all remaining samples. Multiple checks and feedback are incorporated throughout the annotation process to ensure a high-quality outcome.

\subsection{Data Annotation}

\textbf{Question Selection.} 
The question selection process involves the creation of pertinent natural language questions crucial for analyzing new articles within the realm of epidemiology, which were conducted collaboratively with esteemed experts possessing Ph.D. qualifications in the fields of epidemiology and public health. The expertise and knowledge of these professionals contribute to the meticulous design of the questions, which are subsequently categorized into three groups: event questions, epidemiology questions, and ethics questions. Event questions are designed to capture specific information such as the location and disease name, resembling similar disease outbreak events found in \citet{torres2022global} and \citet{carlson2023world}. Epidemiology questions focus on detailed epidemiological information, such as whether the disease was intentionally released, which extends the questions in \citet{conway2009classifying} and \citet{conway2010developing}. Ethics questions require the annotators to conduct thorough ethical assessments to prevent potential privacy breaches or inappropriate content. For a detailed list of these questions, please refer to \Cref{tab:questions}.

\begin{table*}[]
    \centering
    \scriptsize
    \resizebox{1.\textwidth}{!}{
    \begin{tabular}{@{~}L{8cm}@{~}@{~}l@{~}@{~}l@{~}@{~}c@{~}@{~}c@{~}}
\toprule
\textbf{Questions} &             \textbf{Short name} &     \textbf{Category} & \textbf{Options} & \textbf{Sparse}\\
\midrule
1)  Which infectious disease caused the outbreak? &           Disease & Event & -  & -\\
2)  In which country is the outbreak taking place? &           Country & Event & -  & -\\
3)  In which province is the outbreak taking place? &          Province & Event & -  & -\\
4)  In which city/town is the outbreak taking place? &              City & Event & -  & - \\
5)  Check and fill country Geo Code (e.g. 1794299): &       Countrycode & Event & - & -\\
6)  Check and fill province Geo Code (e.g. 1794299): &      Provincecode & Event & -  & -\\
7) Check and fill city Geo Code (e.g. 1815286): &          Citycode & Event & -  & -\\
8)  Which virus or bacteria caused the outbreak? &             Virus & Event & -  & -\\
9)  What symptoms were experienced by the infected victims?  &          Symptoms & Epidemiology & -  & - \\
10) Which institution reported this outbreak? &          Reporter & Epidemiology & -  & - \\
11) What is the type of the victims? &        Victimtype & Epidemiology & Human/Animal/Plant  & -\\
12)  How many new infected cases are reported in the specific event in the report? (please input digits like 1, 34, etc.) &          Casesnum & Epidemiology & -  & -\\
13) Has the victim of the disease travelled across international borders? &           Borders & Epidemiology & YES/NO/Cannot Infer  & YES \\
14) Does the outbreak involve the intentful release? &         Intentful & Epidemiology & YES/NO/Cannot Infer  & YES \\
15) Did human victims acquire the infectious disease from an animal? &        Fromanimal & Epidemiology & YES/NO/Cannot Infer/Not Applicable  & - \\
16) Did the victim fail to respond to a drug? &          Faildrug & Epidemiology & YES/NO/Cannot Infer/Not Applicable  & - \\
17) Are healthcare workers included in the infected victims? & Healthcareworkers & Epidemiology & YES/NO/Cannot Infer  & YES \\
18) Are animal workers included in the infected victims? &     Animalworkers & Epidemiology & YES/NO/Cannot Infer  & YES \\
19) Is the victim of the disease a military worker? &   Militaryworkers & Epidemiology & YES/NO/Cannot Infer  & YES \\
20) Did the outbreak involve a suspected contaminated blood product or vaccine? &           Vaccine & Epidemiology & YES/NO/Cannot Infer  & YES \\
21) Are the victims in a group in time and place? &             Group & Epidemiology & YES/NO/Cannot Infer/Not Applicable  & - \\
22) Did the victim catch the disease during a hospital stay? &      Hospitalstay & Epidemiology & YES/NO/Cannot Infer  & YES \\
23) Is the victim of the disease a child? &             Child & Epidemiology & YES/NO/Cannot Infer  & - \\
24) Is the victim of the disease an elderly person? &           Elderly & Epidemiology & YES/NO/Cannot Infer  & - \\
25) Is the victim of the disease a pregnant woman? &          Pregnant & Epidemiology & YES/NO/Cannot Infer  & YES \\
26) Has the victim of the disease been in quarantine? &        Quarantine & Epidemiology & YES/NO/Cannot Infer  & YES \\
27) Did the outbreak take place during a major sporting or cultural event? &             Event & Epidemiology & YES/NO/Cannot Infer  & YES \\
28) Did the outbreak take place after a natural disaster? &          Disaster & Epidemiology & YES/NO/Cannot Infer  & YES \\
29) When did the outbreak happen? (Relative to article completion time) &             Tense & Epidemiology & Past/Now/Not Yet  & - \\
30) Does the text contain information that can uniquely identify individual people? e.g. names, email, phone, and credit card numbers, addresses, user names. &         Sensitive & Ethics & YES/NO & - \\
\bottomrule
\end{tabular}
}
    \caption{Epidemiology questions given by experts in epidemiology.}
    \label{tab:questions}
\end{table*}

\begin{figure}[t]
    \centering
    \includegraphics[width=0.3\textwidth]{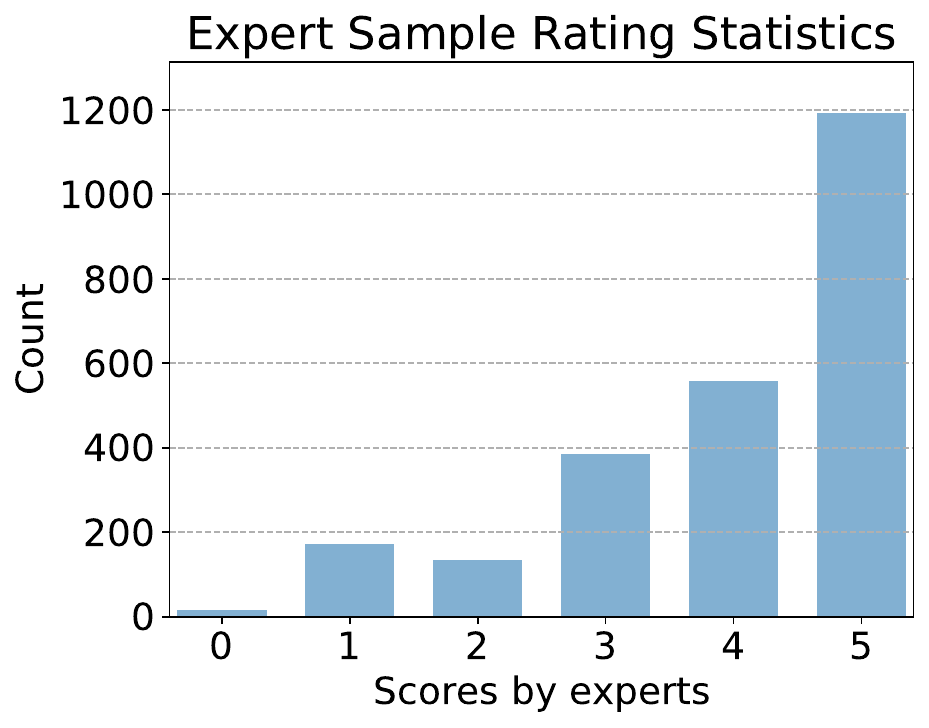}
    \caption{The scores assigned by experts to evaluate the quality of the samples.}
    \label{fig:sampleselection}
\end{figure}

\textbf{Samples Selection.} 
We obtain our raw news alerts from ProMED-mail\footnote{\url{https://promedmail.org/}}, a network of medical professional known for delivering timely information on global disease outbreaks. ProMED-mail covers a wide range of diseases, including infectious diseases, foodborne illnesses, zoonotic diseases, and etc. It offers detailed reports on outbreaks containing crucial information such as the number of cases, outbreak locations, associated symptoms, and etc. This information is invaluable for the development of effective strategies aimed at controlling and preventing the spread of diseases. To conduct our research, we initially collect 36,788 raw alerts available on ProMED-mail, spanning from December 2009 to December 2021.
Then, we engage experts with Ph.D. degrees in epidemiology and public health to generate a list of questions and filter samples for further annotation. Specifically, we carefully select 2,458 samples and request the experts to assign scores ranging from 1 to 5 to each sample. The distribution of scores is illustrated in \Cref{fig:sampleselection}. Samples with scores exceeding 4 are chosen as candidate samples for further analysis. Additionally, it has been observed that certain questions, such as Question 14 (Does the outbreak involve intentful release?), have a sparse distribution of answers, as most diseases are not intentionally released. These questions are called ``sparse questions'' and are listed in \Cref{tab:questions}. To ensure an adequate number of data points for these types of questions, the candidate sample set is ranked based on both the expert scoring and the keyword hits. For instance, if a sample contains keywords like ``intentful release'', the sample will be given one extra point. A detailed list of keywords can be found in Appendix \Cref{tab:keywords}. In this way, samples with more keyword hits are prioritized. However, these kinds of samples are still not enough and a manual search is conducted on ProMED, Wikipedia, and media news platforms to identify relevant articles containing positive answers to these questions.

\textbf{Annotation.} To facilitate the annotation process, we develop a new annotation interface using LabelStudio\footnote{\url{https://labelstud.io/}} (see  \Cref{fig:interface} of Appendix for the annotation interface). Subsequently, we employ a professional annotation company to undertake the annotation of the detailed questions. The annotation process is divided into four batches, comprising 40, 710, 110, and 660 samples, respectively. After each stage, we manually review the annotations and provide feedback to address any systematic annotation issues that may have arisen.

\textbf{Consistency Check.} To ensure the annotation team delivers high-quality annotations, we conducted a quality check during the preliminary annotation. All five annotators were assigned to annotate the same set of 40 samples, and we manually reviewed the answers to identify any obvious mistakes. Additionally, we assessed the consistency of annotations by comparing the responses from all annotators. The comparison results are presented in Appendix \Cref{tab:consistency}. The table demonstrates a high level of consistency among the annotators, which further validates their qualifications for the annotation task.

\textbf{Quality Check.} To maintain the quality of annotations, we implement a quality check after the completion of each batch by the annotators. First, the annotators conduct a manual review of their annotation results to identify and rectify any typos or erroneous annotations. Subsequently, they submit the annotated batch to the experts, who provide feedback to address any misunderstandings that occur. This iterative feedback loop between the annotators and experts ensures ongoing refinement and enhancement of the annotation quality.

\textbf{Ethics Check.} To ensure compliance with ethical requirements, we initiate a research ethics review and obtain permission from the faculty's research ethics committee prior to conducting the annotation process. During the annotation phase, we instruct the annotators to carefully assess whether the samples violate any ethical rules. Any samples found to be in violation were promptly removed from the corpus without further annotation. This proactive approach ensures that the annotation process adheres to ethical guidelines and maintains the integrity of the research.

\subsection{Statistics}
To gain a comprehensive understanding of the BAND dataset, we present various statistics that provide insights into the data's coverage and highlight its significant contributions to the field of NLP and epidemiology. These statistics effectively demonstrate the breadth and depth of the dataset, showcasing its value and potential impact.

\begin{figure*}[t]
    \centering
    \includegraphics[width=1.0\textwidth]{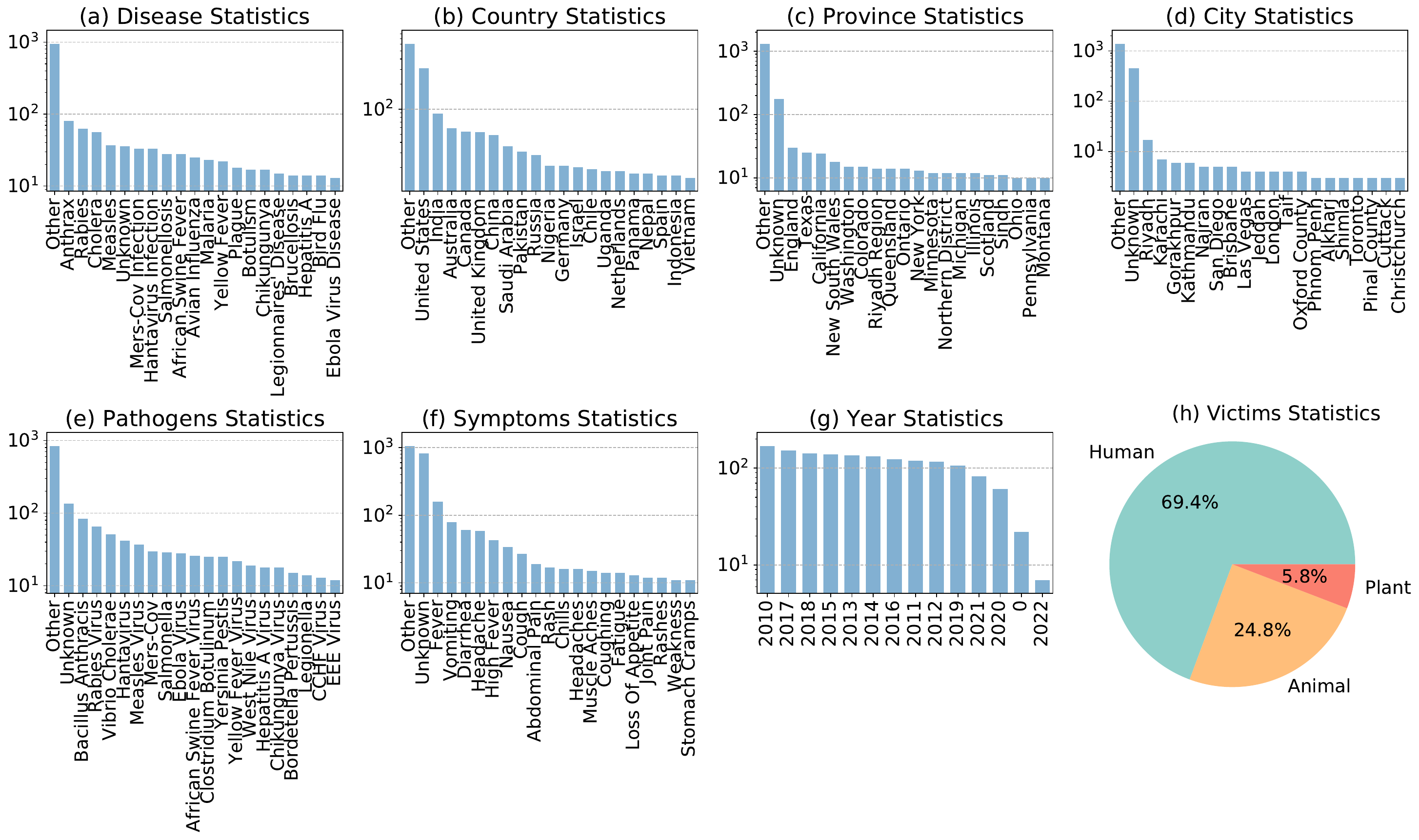}
    \caption{Statistics for BAND corpus}
    \label{fig:stats}
\end{figure*}

\textbf{Disease Distribution.} The distribution of diseases in the BAND dataset is depicted through a histogram shown in \Cref{fig:stats} (a). This histogram reveals that our dataset covers a wide range of popular infectious diseases, such as Anthrax, Cholera, and others. This extensive coverage underscores the dataset's potential for training models to effectively monitor and surveil various infectious disease outbreaks.

\textbf{Location Distribution.} In addition, we have generated visualizations of the location distribution in the BAND dataset, highlighting the coverage of various countries (\Cref{fig:stats} (b)), provinces (\Cref{fig:stats} (c)), and cities (\Cref{fig:stats} (d)). These visualizations demonstrate that our dataset encompasses a wide range of locations, affirming its potential for training a model capable of handling daily news reports from around the world. This global coverage further enhances the dataset's applicability in addressing diverse epidemiological scenarios.

\textbf{Pathogen Distribution.} Our dataset exhibits a comprehensive coverage of various  pathogens including bacteria, fungi, protozoa, viruses, and etc. The distribution is shown in \Cref{fig:stats} (e). It is evident from the statistics that the BAND dataset encompasses mentions of numerous popular infectious pathogens, including bacillus anthracis, rabies virus, vibrio cholera, and many others. This extensive coverage of prominent pathogens enhances the dataset's relevance and suitability for training models to effectively analyze and respond to a wide range of infectious disease scenarios.

\textbf{Victim Distribution.}
Within the BAND dataset, the term ``victim'' refers to the infected host type, which includes humans, animals, and plants. As depicted in \Cref{fig:stats} (h), our primary focus is on human and animal diseases. However, we have also included a portion of the data (approximately 6\%) that describes plant diseases, thereby extending the application domains of the dataset.

\textbf{Symptoms Distribution.} The BAND dataset includes annotations for a diverse range of symptoms, as illustrated in \Cref{fig:stats} (f). Symptoms such as fever, vomiting, and others are comprehensively covered within the dataset. This broad coverage of symptoms highlights the potential to train a model capable of effectively handling different types of symptoms.

\subsection{Data Split}

We provide two different sampled splits, namely the Rand Split and the Stratified Split, for our BAND corpus, as shown in \Cref{tab:datasplit}.

\textbf{Rand Split.} This split randomly partitions the corpus into train/dev/test sets, without considering any other factors or criteria.

\textbf{Stratified Split.} In order to assess the model's ability to accurately answer sparse questions with limited positive answers, it is crucial to focus on these specific samples in upcoming research. To accomplish this, we employ a split strategy that prioritizes samples with positive answers for sparse questions. These samples are divided in a ratio of 5:1:4 for the train/dev/test sets respectively. This ensures that these important samples are adequately trained and evaluated. Then, we randomly sample other instances to complement the dataset. This approach allows for a thorough assessment of the model's performance in addressing the challenges posed by sparse questions.

\begin{table}[]
    \centering
    \scriptsize
    \begin{tabular}{lrr}
        \toprule
        {} &  \textbf{Rand} &  \textbf{Stratified} \\
        \midrule
        train &  1,208 &        1,126 \\
        dev   &   150 &         149 \\
        test  &   150 &         233 \\
        \bottomrule
    \end{tabular}

\caption{Data split.}
    \label{tab:datasplit}
\end{table}

\section{Experiments}
To evaluate the performance of existing NLP models on our newly annotated dataset, we conduct experiments on three widely used NLP tasks: Named Entity Recognition (NER), Question Answering (QA), and Event Extraction (EE). By assessing the performance of various models on these tasks, we can gain insights into their strengths and limitations in handling this dataset.

\subsection{Experimental Setup}

\subsubsection{NER Task} The NER task aims at extracting named entities belonging to specific categories. In this study, we aim to demonstrate how our annotated biomedical dataset can help advance research in the NER task for specific terms. We focus on extracting entities related to disease names, outbreak locations (country/province/city), pathogens (viruses/bacteria), and symptoms. To ensure compliance with the requirements of the NER task, we limit our analysis to only those entities that are explicitly mentioned in the original text, without considering any additional inference by annotators. We compare the performance of various models, including CRFBased, TokenBased, SpanBased, and ChatGPT models in extracting named entities. The model details are as follows:

\textbf{CRFBased} model \cite{lafferty2001conditional,mccallum2003early,manning2014stanford} incorporates the contextual information of nearby words to recognize and classify named entities in the provided text. It utilizes a Conditional Random Field (CRF) layer to predict the BIO tags for each input word.

\textbf{TokenBased} model \cite{kenton2019bert,lee2020biobert} fine-tunes a pre-trained model on the annotated NER data and uses it to directly predict the BIO tags for each word in the given sequence.

\textbf{SpanBased} model \cite{lee2017end,luan2018multi,luan2019general,wadden2019entity,zhong2021frustratingly} partitions the input text into spans of varying lengths and then directly assigns labels to the spans that correspond to named entities. This technique has been demonstrated to improve the performance of previous NER models. We employ the implementation provided by \citet{zhong2021frustratingly}.

\textbf{ChatGPT} model \cite{ouyang2022training} has shown great potential in performing this task without requiring additional training or fine-tuning. We elicit named entities with corresponding categories directly from the API. We have attempted various prompts to instruct ChatGPT to infer as much information as possible and adhere to the terminology mentioned in the original text.

\subsubsection{QA Task} The task of question answering involves providing answers to questions based on a given corpus. This task can be categorized as either extractive QA or abstractive QA. In extractive QA, the model selects the relevant answer span from the input text, while in abstractive QA, the model generates an answer based on the input text, which may not be an exact span from the given text. In our task, as the answer to many questions may not exist in the original text, we focus on the abstractive QA setting. We demonstrate the performance of several models to showcase their potential on our dataset. To evaluate the models' performance, we utilize the widely-used accuracy metric. Prior to comparing the models' results to the gold standard label, we normalize all occurrences of ``N/A'', ``Unknown'', ``na'', and ``nan'' as ``Cannot Infer''. We use exact match accuracy to evaluate all the results. We conduct our experiments based on the following widely used models:

\textbf{T5} \cite{raffel2020exploring} model is built on the Transformer architecture and is pre-trained on large volumes of text data using a diverse range of language modeling tasks. We fine-tune the T5 model on the training set by concatenating the text and the question as input, with the output being the answer to the corresponding question.

\textbf{Bart} \cite{lewis2019bart} model is similar to the T5 model, as it also employs an encoder-decoder architecture. We use the Bart model as the backbone model and fine-tune it on our annotated dataset using the same setting as the T5 model.

\textbf{GPT2} \cite{li2021prefix}  is a decoder-only language model that concatenates all context sequences, questions, and answers into a single sequence, which is then used to fine-tune the GPT2 model.

\textbf{GPTNEO} \cite{black2022gpt} is a transformer-based language model developed by EleutherAI, designed to be an open-source model similar to GPT-3. It was trained on the Pile dataset, which comprises a diverse corpus of text data, including books, websites, and academic papers.

\textbf{OPT} \cite{zhang2022opt} is a decoder-only language model developed by Meta AI, with the aim of providing an open-source model comparable to GPT-3. OPT offers models with parameters ranging from 125M to 175B. In our experiment, we adopt the model with 350M parameters.

\textbf{Galactica} \cite{taylor2022galactica} is a decoder-only language model trained on a large-scale scientific corpus. It is designed to handle scientific tasks, including scientific QA, mathematical reasoning, summarization, and document generation. The model may have been trained with corresponding disease and country names, making it more likely to understand the news text in our dataset. Galactica comes in a range of model sizes, from 125M to 120B parameters, and we test the 125M model in our experiment as larger models tend to explain the answer with their own words instead of our pre-defined format, leading to a degenerated performance.

\textbf{BLOOM} \cite{scao2022bloom}  is an autoregressive large language model that outputs coherent text in 46 languages and 13 programming languages. Additionally, it can complete diverse text tasks, even those it was not directly trained for, by framing them as text generation tasks.

\textbf{ChatGPT} \cite{ouyang2022training} model is a zero-shot model that cannot be fine-tuned using our training set. We use its API similar to how we use it for the NER task. Firstly, we prompt ChatGPT to read the paragraph and then ask it to answer the related questions one by one. The instructions for ChatGPT can be found in Appendix \Cref{fig:chatinstruct}.

\subsubsection{EE Task} 
The primary goal of event extraction is to automatically identify and extract relevant information from unstructured text data. The identification of disease outbreak events can be challenging due to the use of diverse terminology. Although triggers like ``outbreak'', ``epidemic'', or ``pandemic'' may be utilized, their absence can limit the effectiveness of traditional keyword-based approaches. In response, we defined a set of attributes for outbreak events: disease name, location, pathogens involved, victim type, and associated symptoms. Our task involves extracting these attributes from a document as input, employing autoregressive models that are similar to the ones utilized in our question-answering baselines.

% advantages of generation-based methods:
%     \item tackle multi-label situations, one argument has multiple roles
%     \item extract attributes that are not exact match of words
% \end{itemize}

\subsection{Experimental Results}

\begin{table}[h]
    \centering
    \scriptsize
    \begin{tabular}{@{~}l@{~}@{~}r@{~}@{~}r@{~}@{~}r@{~}@{~}r@{~}@{~}r@{~}@{~}r@{~}}
        \toprule
        & \multicolumn{3}{c}{\textbf{Random}} & \multicolumn{3}{c}{\textbf{Stratified}} \\
        \cmidrule(lr){2-4} \cmidrule(lr){5-7}
        \textbf{Model} & \textbf{Precision} & \textbf{Recall} & \textbf{F1-score} & \textbf{Precision} & \textbf{Recall} & \textbf{F1-score} \\
        \midrule
        CRFBased & 0.582 & 0.674 & 0.625 & 0.600 & 0.663 & 0.630 \\
        TokenBased & 0.631 & 0.691 & 0.660 & 0.701 & 0.730 & 0.715 \\
        SpanBased & 0.598 & 0.694 & 0.642 & 0.676 & 0.759 & 0.715 \\
        ChatGPT & 0.326 & 0.353 & 0.339 & 0.424 & 0.318 & 0.363 \\
        \bottomrule
    \end{tabular}
    \caption{Named entity recognition results.}
    \label{tab:nermain}
\end{table}

\begin{table}[]
    \centering
    \scriptsize
\begin{tabular}{lrrr}
\toprule
{} &      \textbf{Precision} &      \textbf{Recall} &     \textbf{F1-score} \\
\midrule
City     &  0.326 &  0.500 &  0.395 \\
Country  &  0.710 &  0.760 &  0.734 \\
Disease  &  0.583 &  0.758 &  0.659 \\
Province &  0.616 &  0.517 &  0.562 \\
Virus    &  0.696 &  0.823 &  0.754 \\
\bottomrule
\end{tabular}
\caption{NER results for each domain.}
    \label{tab:nerdomain}
\end{table}

\textbf{NER Task.} The results for NER task are shown in \Cref{tab:nermain}. The results indicate that: 1) The existing supervised models (CRFBased, TokenBased, SpanBased) achieve good performance than the zero-shot model (ChatGPT), which suggests that training the model with the BAND corpus can aid in identifying commonly used named entities in disease outbreak news. 2) ChatGPT does not perform as well as the supervised models. This could be due to several factors: firstly, our data is newly annotated and belongs to a highly specialized domain that ChatGPT may not have been extensively trained on. Additionally, ChatGPT prefers to use its own words to provide the name (which is usually more formal), leading to lower scores. We have attempted to utilize multiple instructions to encourage it to use the original text (as shown in \Cref{fig:chatinstruct}), but it remains unresponsive.

We also show the NER results for each domain in \Cref{tab:nerdomain}. The following observations can be made: 1) The NER model performs well in the country, disease, and virus domains. This is likely because the named entities in the testing set are also present in the training set, and the model has learned to recognize these types of entities. 2) In the province and city domains, there is a significant drop in the F1 score because some of the city or province names are not mentioned in the training set. As a result, our dataset necessitates better few-shot/zero-shot capabilities of the model, which presents new challenges to this field.

\begin{table}[ht]
    \centering
    \scriptsize
    \setlength{\tabcolsep}{5pt}
    \begin{tabular}{lrrll}
    \toprule
    \textbf{Model} & \textbf{Rand} & \textbf{Stratified} & \textbf{Size} & \textbf{Mode} \\
    \midrule
    T5 & 0.674 & 0.591 & 220M (base) & Finetune \\
    Bart & 0.666 & 0.510 & 140M (base) & Finetune \\
    \hline
    GPT2 & 0.663 & 0.647 & 124M & Finetune \\
    OPT & 0.699 & 0.687 & 125M & Finetune \\
    GPTNEO & 0.695 & 0.695 & 125M & Finetune \\
    Galactica & 0.717 & 0.710 & 125M & Finetune \\
    BLOOM & 0.735 & 0.751 & 560M & Finetune \\
    \hline
    ChatGPT & 0.497 & 0.413 & - & Zero-Shot \\
    \bottomrule
    \end{tabular}
    \caption{Question answering results.}
    \label{tab:qaresults}
    \end{table}

\begin{table*}[]
    \centering
    \scriptsize
    \resizebox{1.\textwidth}{!}{
    \begin{tabular}{@{~}l@{~}@{~}l@{~}|@{~}L{1.5cm}@{~}l@{~}@{~}l@{~}@{~}l@{~}@{~}l@{~}@{~}l@{~}@{~}l@{~}@{~}l@{~}@{~}l@{}}
    \toprule
    \textbf{Model} & {} & \textbf{2) Country} & \textbf{3) Province} & \textbf{13) Borders} & \textbf{14) Intentful} & \textbf{15) Fromanimal} & \textbf{18) Animalworkers} & \textbf{23) Child} & \textbf{25) Pregnant} & \textbf{28) Disaster} \\
    \hline
    T5 & Accuracy & 0.78 & 0.52 & 0.7 & 0.927 & 0.72 & 0.787 & 0.767 & 0.847 & 0.92 \\
    & Predict & Ukraine & -- & Cannot Infer & -- & NO & NO & NO & NO & -- \\
    & Gold Standard & Russia & Ohio & NO & NO & Cannot Infer & Cannot Infer & Cannot Infer & Cannot Infer & NO \\
    & Error Count & 2 & 1 & 30 & 1 & 14 & 10 & 7 & 9 & 1 \\ \hline
    BLOOM & Accuracy & 0.794 & 0.442 & 0.833 & 0.97 & 0.781 & 0.82 & 0.824 & 0.807 & 0.961 \\
    & Predict & DR Congo & nan & Cannot Infer & NO & NO & Cannot Infer & NO & NO & NO \\
    & Gold Standard & Democratic Republic of Congo & Helmand & YES & YES & Cannot Infer & NO & Cannot Infer & Cannot Infer & YES \\
    & Error Count & 2 & 1 & 25 & 6 & 25 & 22 & 14 & 21 & 8 \\ \hline
    ChatGPT & Accuracy & 0.403 & 0.236 & 0.678 & 0.056 & 0.176 & 0.519 & 0.489 & 0.528 & 0.361 \\
    & Predict & Cannot Infer & Cannot Infer & No & Cannot Infer & Cannot Infer & Cannot Infer & Cannot Infer & Cannot Infer & Cannot Infer \\
    & Gold Standard & United States & California & Cannot Infer & NO & NO & NO & NO & NO & NO \\
    & Error Count & 37 & 5 & 50 & 219 & 104 & 62 & 106 & 105 & 143 \\
    \bottomrule
    \end{tabular}
    }
    \caption{Error analysis for accuracy of each question and top 1 error statistics for T5, BLOOM, and ChatGPT models.}
    \label{tab:chatgpterror}
\end{table*}

\textbf{QA Task.} The results of QA task are shown in \Cref{tab:qaresults}. It can be observed from the results that 1) the decoder-only framework (e.g. GPT2 and Galactica) generally outperforms the encoder-decoder framework (e.g. T5 and Bart). This could be because decoder-only models are pre-trained on more extensive text data. 2) The BLOOM model outperforms other generative models, which may be due to its training on corpora more relevant to our domain. Additionally, finetuning a larger BLOOM model can force it to use our desired output style, while other larger backbone models tend to use their own words to answer. 3) We also attempt to utilize ChatGPT, but its performance was not as good as fine-tuned models. This could be because ChatGPT is a zero-shot model that has not been extensively trained on our new dataset. Additionally, even after utilizing various instructions to encourage inference, ChatGPT did not perform any inference. In the next subsection, we conduct a detailed error analysis to discuss these issues.

\textbf{Error Analysis.} We conduct an error analysis on three types of models by selecting questions with lower accuracy and analyzing their performance, as present in \Cref{tab:chatgpterror}. Our observations reveal that ChatGPT faces challenges in determining when to conduct inference or simply answer ``Cannot Infer''. There are instances where ChatGPT refrains from making any inference, even when a straightforward inference could yield the correct answer. This suggests that ChatGPT may have limitations in certain cases, despite attempts to encourage inference through various instructions. For example, in one scenario involving a disease outbreak in Los Angeles, ChatGPT correctly identify the outbreak city but consistently fail to infer the country, possibly due to internal restrictions in its question-answering task that focus solely on the provided text, disregarding its background knowledge for inference. Moreover, in certain questions (e.g., Question 13), ChatGPT erroneously infers information, whereas human annotators tend to avoid such inferences. These findings underscore the ongoing discrepancy between ChatGPT and human experts in determining when information can be inferred.

\textbf{EE Task.} The results of event extraction (EE) are presented in \Cref{tab:eemain}, which includes both overall and class-specific F1 scores to assess the performance. A common trend observed in all models is their inability to perform well in certain categories such as ``province code'' and ``city code'', with several models scoring zero. While BART stands out as a superior performer in the geocode prediction task, it suffers from a loss of precision when it comes to extracting other context-related attributes. Remarkably, ChatGPT outperforms fine-tuned decoder-only models in the zero-shot setting, showcasing its ability to extract latent knowledge from pre-training data beyond the context at hand. Our study highlights the varied strengths of different models across multiple categories. Notably, in the event extraction task, encoder-decoder models, such as T5 and BART, exhibit significantly better performance than decoder-only models. This finding is different from the results in the QA task and may suggest a deficiency in the ability of decoder-only models to generate structured text compared to encoder-decoder models. This is because, in the QA task, There are a lot of YES/NO questions to be answered which do not exist in the event extraction task. 

%For this task, we opted to utilize the same text generation models that served as baselines for our question answering evaluations. As we previously discussed, several outbreak attributes targeted for extraction are not presented in a canonical form within the source text. Further, certain attributes, such as the country geocode and victim type, are not explicitly provided in the context. By treating the event extraction task as a sequence generation task, we are able to leverage the inherent knowledge and inference capabilities of the large language models (LLMs). 

%From our evaluation, we found that T5 showcased the highest overall performance, with an F1 score of 60.88. Furthermore, this model also excelled in the "disease" category, achieving an F1 score of 76.41, and in the "symptoms" category, where it achieved a score of 76.75. Bart was a close second in overall performance, with an F1 score of 60.86. This model demonstrated exceptional performance in the "country" and "country code" classes, with scores of 88.16 and 85.15, respectively. GPT2 and OPT presented comparable overall performances, with F1 scores of 45.34 and 48.27 respectively. However, GPTNEO exhibited the lowest overall performance with an F1 score of 34.34. 

\begin{table*}[htbp]
    \centering
    \scriptsize
    \resizebox{1.\textwidth}{!}{
  \begin{tabular}{@{~}c@{~}@{~}c@{~}@{~}c@{~}@{~}c@{~}@{~}c@{~}@{~}c@{~}@{~}c@{~}@{~}c@{~}@{~}c@{~}@{~}c@{~}@{~}c@{~}@{~}c@{~}}
    \hline
    \textbf{Model} & \textbf{Overall F1} & \multicolumn{7}{c}{\textbf{Individual F1}} \\
    \cline{3-12}
     &  & \textbf{Disease} & \textbf{Country} & \textbf{Province} & \textbf{City}  & \textbf{Country code} & \textbf{Province code} & \textbf{City code} & \textbf{Pathogen} & \textbf{Symptoms} & \textbf{Victim} \\
    \hline
    T5 & 60.88 & 76.41 & 87.87 & 56.44 & 58.02 & 68.63 & 15.05 & 2.33 & 66.17 & 76.75 & 97.33 \\
    Bart & 60.86 & 68.29 & 88.16 & 49.52 & 53.28 & 85.15 & 32.81 & 6.64 & 53.58 & 61.54 & 98.67 \\
    GPT2 & 45.34 & 63.92 & 78.43 & 38.24 & 34.48 & 38.69 & 0 & 0 & 49.82 & 41.86 & 93.65\\
    OPT & 48.27 & 66.89 & 82.51 & 49.54 & 43.09 & 32.24 & 0.62 & 0 & 52.99 & 54.68 & 94.31\\
    GPTNEO & 34.34 & 58.42 & 62.0 & 31.97 & 30.49 & 18.24 & 0 & 0 & 31.73 & 19.13 & 74.05\\
    Galactica & 49.33 & 62.35 & 78.95 & 50.33 & 45.97 & 46.05 & 0.65 & 0 & 56.72 & 57.61 & 91.47\\
    Bloom & 48.40 & 62.05 & 78.29 & 40.0 & 49.81 & 48.68 & 1.96 & 0 & 48.89 & 53.54 & 94.28\\
    ChatGPT & 47.71 & 56.16 & 79.15 & 47.29 & 46.15 & 51.61 & 7.32 & 4.4 & 28.04 & 45.03 & 83.5\\
    % T5-stratified & 74.26 & 80.09 & 85.41 & 58.37 & 63.11 & & & & 66.97 & 68.09 & 98.49 \\
    % Bart-stratified & 67.83 & 77.78 & 87.98 & 56.8 & 55.58 & & & & 40.84 & 58.48 & 97.63\\
    \hline
  \end{tabular}
  }
\caption{Event extraction results on random split.}
\label{tab:eemain}
\end{table*}

\section{Related Works}
Numerous epidemiology disease surveillance systems have been developed to monitor disease outbreak events. Among these, the BioCaster system \cite{meng2022biocaster} automatically gathers news and alerts from social media, while GPHIN \cite{mawudeku2013global} uses global surveillance and data analysis to detect potential public health threats. ProMED-mail \cite{yu2004promed} relies on a network of experts to provide real-time news alerts and expert notifications on emerging diseases and outbreaks. HealthMap \cite{freifeld2008healthmap} aggregates disease data from various sources to provide real-time disease outbreak monitoring and visualization. FluTrackers \footnote{\url{https://flutrackers.com/forum/}} offers real-time monitoring and analysis of influenza. The ECDPC \footnote{\url{https://www.ecdc.europa.eu/en}} is utilized for real-time monitoring, risk assessment, and outbreak investigation of infectious diseases in Europe. However, these systems have only made use of simple NLP tools such as extraction tools, and further data is needed to train models with a deeper understanding of news articles and reports.

In order to enhance the performance of disease surveillance systems, several pandemic and epidemic datasets have been annotated. \citet{conway2010developing} have annotated a disease outbreak dataset comprising 200 samples. Meanwhile, \citet{torres2022global} have provided a dataset that includes 2,227 samples; however, this dataset mainly concentrates on the outbreak event itself and does not contain report text annotations. \citet{chan2010global,carlson2023world} have utilized WHO's data to examine the outbreak event. \citet{mutuvi2020dataset,mutuvi2020multilingual} have emphasized multilingual outbreak detection, whereas \citet{balashankar2019reconstructing} and \citet{lamsal2021design} have centered on outbreak news for MERS and COVID-19, respectively. Nevertheless, these datasets are either relatively small or emphasize outbreak statistics ignoring other important information for epidemiologists.

\section{Conclusions}
In this paper, we contribute a new Biomedical Alert News Dataset (BAND) designed to provide a more comprehensive understanding of disease spread and epidemiology related questions by enabling NLP systems to analyze and answer several important questions. Our dataset contains 1,508 samples from recent news articles, open emails, and alerts, as well as 30 event and epidemiology related questions. The questions and samples are carefully selected by domain experts in the fields of epidemiology and NLP and require common sense reasoning capability of NLP models. BAND is the largest corpus of well-annotated biomedical alert news with elaborately designed questions, and we provide a variety of model benchmarks for Named Entity Recognition (NER), Question Answering (QA), and Event Extraction (EE) in the epidemiology domain. The experimental results show that the new dataset can help train NLP models to better understand disease outbreak events and answer important epidemiology questions.

\section{Limitations}

While the BAND corpus provides valuable insights into disease outbreak events and enhances understanding in the field of epidemiology, it has several limitations. First, the dataset primarily focuses on disease outbreak news reports obtained from ProMED-mail, which may introduce bias towards certain types of outbreaks or regions. This limited scope may not capture the full diversity of disease outbreaks and their characteristics. Second, the annotation process is subjective and relies on the expertise and judgments of human annotators, which introduces potential biases and inconsistencies in the annotations. Despite efforts to ensure inter-annotator agreement, variations in interpretation and annotation errors are still possible. Finally, the dataset predominantly focuses on English-language news articles, limiting its generalizability to other languages and regions where news reports may be in different languages.

\section{Broader Impact Statement}

The BAND corpus and its associated research, after successfully passing an ethics panel review and implementing ethical checks in the annotation process. By ensuring ethical standards are upheld, including fair compensation for annotators, this project promotes responsible research practices. Moreover, it is important to note that all the raw corpus used in this study is sourced from publicly available internet data, with no utilization of private or sensitive information. Therefore, it ensures data privacy and mitigates any potential concerns regarding the use of personal data, further establishing the ethical integrity of the project. 

\section*{Acknowledgments}
The authors gratefully acknowledge the support of the funding from UKRI under project code ES/T012277/1. We would also like to express our sincere gratitude to Anya Okhmatovskaia and Nicholas King for their invaluable assistance and insightful discussions.

\bibliography{reference}
\bibliographystyle{acl_natbib}

\clearpage
\appendix
\onecolumn

\begin{center}
    {\LARGE \textbf{Appendix. Supplementary Material}}
\end{center}

\section{Keywords for Questions}
\label{sec:appendix}

The keywords for each samples are shown in \Cref{tab:keywords}. In order to provide enough data points for sparse questions, we rank the candidate sample set according to keywords hit. For example, if the text contains keywords ``intentful release'', we add one score for the sample. We choose samples with more keywords hits. Unfortunately, these kinds of samples are still not enough, we manually search the relevant keywords on Promed, Wikipedia and the media news and add articles containing sparse answers to these questions into the sample list.

\begin{table*}[h]
    \centering
    \scriptsize

\begin{tabular}{L{5cm}L{9cm}r}
    \toprule
                                                                           question &                                                                                                                                                                                                                                                                                                                                                                                                                                                                                                                                                                Keywords &  Count \\
    \midrule
          11) Has the victim of the disease travelled across international borders? &                                                                                                                                                                          international, Flight, flight, airline, airport, travel, sightseeing, tour, travel, vacation, visit, air, airline, airlines, airway, aviation, cargo, charter, crew, flier, fliers, passenger, tourist, traveler, traveller, travellers, crew member, sea crew member, air crew member, airline crew member, sea crew, airline crew, air crew, seaman, sea men, seamen, migrant, border migrant &   1611 \\
                               12) Does the outbreak involve the intentful release? &                                                                                                                                                                                                                                                                                                                                                                                                                                                                                       intentful release, terror, white powder, extremist,  subversive &     12 \\
                       15) Are healthcare workers included in the infected victims? &                                                                                                                                                                                                                                             healthcare worker, clinic, careworker, doctor, health worker, healthcare worker, hospital staff, hospital worker, medic, nurse, physician, psychiatric worker, surgeon, pediatrician, public health worker, GP, general practitioner, lab worker, laboratory worker, health staff, care worker, residential aged care worker &    902 \\
                           16) Are animal workers included in the infected victims? &                                                                                                                                                                                                                                                    butcher, animal worker, shepherd, rancher, veterinary, abattoir worker, animal skin processor, breeder, culler, farmer, forester, keeper, livestock handler, livestock market employee, peasant, poultry, rancher, slaughterhouse, slaughterman, slaughtermen, trader, transporter, vet, veterinarian, animal breeder &   1700 \\
                                17) Is the victim of the disease a military worker? &                                                                                                                                                                                                                                                                                                                                                                                                                                                                                                                military, soldier, defense, defence force, service member &    155 \\
    18) Did the outbreak involve a suspected contaminated blood product or vaccine? &                                                                                                                                                                                                                                                                                                                                                                                                                                                                                                                                                           blood, vaccine &   2294 \\
                                          22) Is the victim of the disease a child? &                                                                                                                                                                                         child, boy, girl, adolescent, child, juvenile, kids, minors, pupil, school children, schoolchildren, student, teenager, youngster, infant, baby, babies, preschool, newborn, pediatric, 1-year-old, 2-year-old, 3-year-old, 4-year-old, 5-year-old, 6-year-old, 7-year-old, 8-year-old, 9-year-old, 10-year-old, 11-year-old, 12-year-old, 13-year-old, 14-year-old, 15-year-old &   1239 \\
                               23) Is the victim of the disease an elderly person?  & retired, older, elderly, grandfather, grandmother, grandparents, care home resident, older people, aged care resident, nursing home resident, elderly parents, older people, elderly patient, 60-year-old, 70-year-old, 80-year-old, 71-year-old, 72-year-old, 73-year-old, 74-year-old, 75-year-old, 76-year-old, 77-year-old, 78-year-old, 79-year-old, 81-year-old, 82-year-old, 83-year-old, 84-year-old, 85-year-old, 86-year-old, 87-year-old, 88-year-old, 89-year-old, 91-year-old, 92-year-old, 93-year-old, 94-year-old, grandson, granddaughter, grandaughter &    564 \\
                                 24) Is the victim of the disease a pregnant woman? &                                                                                                                                                                                                                                                                                                                                                                                                                                                                                                                                               pregnant, expecting mother &    309 \\
                              25) Has the victim of the disease been in quarantine? &                                                                                                                                                                                                                                                                                                                                                                                                                                                                                                                                                               quarantine &    421 \\
         26) Did the outbreak take place during a major sporting or cultural event? &                                                                                                                                                                                                                                                                                                                                                                                                                                                                   sport, olympic, games, tournament, marathon, congress, festival, games, championship, fair, pilgrimage &     94 \\
                          27) Did the outbreak take place after a natural disaster? &                                                                                                                                                                                                                                                                                                                                                                                                                                                                                                                                    earthquake, hurricane, flood, drought &     94 \\
                                               30) What is the type of the victims? &                                                                                                                                                                                                                                                                                                                                                                                                                                                                                                           animal, horse, bird, chicken, cat, dog, pig, deer, plant, tree &   2901 \\
    \bottomrule
    \end{tabular}

    \caption{Keywords}
    \label{tab:keywords}
\end{table*}

\section{Annotation Consistency Check}

\Cref{tab:consistency} shows the results of the annotation consistency check. We ask 5 annotators to annotate the sampe sample and compare the corresponding results and the table provides an overview of various questions and their corresponding statistics related to annotator agreement. 

The first column is the different questions that are posed to the annotators. The subsequent columns provide information on the total number of answers received for each question, the count of cases where all annotators provided the same answer, the count of cases where at least four annotators agreed on the same answer, the count of cases where at least three annotators agreed on the same answer, the number of unique answers received, and the most common answer for each question. Additionally, the table includes information about the portion of responses that the most common answer and the second most common answer account for.

The results show varying levels of agreement among annotators for different questions. For example, in the case of Disease, 6 out of 40 answers had unanimous agreement among annotators, while in the case of Country, 26 out of 40 answers are consistent among all annotators. Similarly, the table provides insights into the most common answers and their portions for each question.

Overall, this annotation consistency check demonstrates the extent of agreement among annotators for different questions in the study. The results will help in evaluating the reliability and consistency of the annotations, providing valuable insights for further analysis and interpretation of the data.

\begin{table*}[]
    \centering
    \scriptsize
    \begin{tabular}{l@{~}C{1.5cm}@{~}@{~}C{1.5cm}@{~}@{~}C{1.5cm}@{~}@{~}C{1.5cm}@{~}@{~}C{1.5cm}@{~}@{~}C{1.5cm}@{~}@{~}C{1.5cm}@{~}@{~}C{1.5cm}@{~}@{~}C{1.5cm}}
\toprule
         Questions &  Total answer &  All annotators give the same answer &  >=4 annotators give the same answer &  >=3 annotators give the same answer &  Unique answer &     Most common answer &  Most common portion &                  Second common answer &  Second common portion \\
\midrule
           Disease &            40 &                                    6 &                                   22 &                                   36 &             70 &             (Anthrax,) &                0.070 &                       (Yellow fever,) &                  0.050 \\
           Country &            40 &                                   26 &                                   35 &                                   39 &             45 &       (United States,) &                0.235 &                             (Canada,) &                  0.050 \\
          Province &            40 &                                   19 &                                   26 &                                   34 &             56 &                        &                0.310 &                             (Kansas,) &                  0.030 \\
              City &            40 &                                   16 &                                   22 &                                   32 &             70 &                        &                0.245 &                  (Washington County,) &                  0.025 \\
       Countrycode &            40 &                                   39 &                                   39 &                                   40 &             30 &             (6252001,) &                0.250 &                            (6251999,) &                  0.050 \\
      Provincecode &            40 &                                   20 &                                   28 &                                   39 &             46 &                        &                0.310 &                            (4273857,) &                  0.030 \\
          Citycode &            40 &                                   24 &                                   31 &                                   39 &             48 &                        &                0.245 &  (2639272, 2641022, 2642593, 6691235) &                  0.025 \\
             Virus &            40 &                                   14 &                                   23 &                                   27 &             85 &  (Bacillus anthracis,) &                0.095 &                 (Yellow fever virus,) &                  0.050 \\
          Symptoms &            40 &                                   27 &                                   28 &                                   32 &             48 &                        &                0.660 &                              (Fever,) &                  0.045 \\
          Reporter &            40 &                                   24 &                                   28 &                                   36 &             39 &                        &                0.510 &                      (Fraser Health,) &                  0.025 \\
        Victimtype &            40 &                                   40 &                                   40 &                                   40 &              3 &               (Human,) &                0.675 &                             (Animal,) &                  0.300 \\
          Casesnum &            40 &                                   32 &                                   39 &                                   40 &             26 &                   (1,) &                0.285 &                                       &                  0.125 \\
           Borders &            40 &                                   32 &                                   37 &                                   40 &              2 &        (Cannot Infer,) &                0.870 &                                 (NO,) &                  0.130 \\
         Intentful &            40 &                                   40 &                                   40 &                                   40 &              1 &                  (NO,) &                1.000 &                                       &                  0.000 \\
        Fromanimal &            40 &                                   39 &                                   40 &                                   40 &              4 &                  (NO,) &                0.370 &                     (Not Applicable,) &                  0.325 \\
          Faildrug &            40 &                                   38 &                                   40 &                                   40 &              4 &        (Cannot Infer,) &                0.540 &                     (Not Applicable,) &                  0.325 \\
 Healthcareworkers &            40 &                                   37 &                                   39 &                                   40 &              3 &        (Cannot Infer,) &                0.580 &                                 (NO,) &                  0.415 \\
     Animalworkers &            40 &                                   37 &                                   40 &                                   40 &              3 &        (Cannot Infer,) &                0.595 &                                 (NO,) &                  0.385 \\
   Militaryworkers &            40 &                                   37 &                                   40 &                                   40 &              2 &        (Cannot Infer,) &                0.610 &                                 (NO,) &                  0.390 \\
           Vaccine &            40 &                                   40 &                                   40 &                                   40 &              1 &                  (NO,) &                1.000 &                                       &                  0.000 \\
             Group &            40 &                                   35 &                                   39 &                                   40 &              4 &        (Cannot Infer,) &                0.615 &                     (Not Applicable,) &                  0.275 \\
      Hospitalstay &            40 &                                   30 &                                   37 &                                   40 &              2 &                  (NO,) &                0.630 &                       (Cannot Infer,) &                  0.370 \\
             Child &            40 &                                   37 &                                   39 &                                   40 &              3 &                  (NO,) &                0.615 &                       (Cannot Infer,) &                  0.240 \\
           Elderly &            40 &                                   37 &                                   40 &                                   40 &              3 &        (Cannot Infer,) &                0.470 &                                 (NO,) &                  0.450 \\
          Pregnant &            40 &                                   37 &                                   40 &                                   40 &              3 &                  (NO,) &                0.485 &                       (Cannot Infer,) &                  0.465 \\
        Quarantine &            40 &                                   35 &                                   39 &                                   40 &              3 &        (Cannot Infer,) &                0.850 &                                 (NO,) &                  0.095 \\
             Event &            40 &                                   29 &                                   38 &                                   40 &              3 &        (Cannot Infer,) &                0.570 &                                 (NO,) &                  0.405 \\
          Disaster &            40 &                                   40 &                                   40 &                                   40 &              1 &                  (NO,) &                1.000 &                                       &                  0.000 \\
             Tense &            40 &                                   40 &                                   40 &                                   40 &              2 &                 (Now,) &                0.925 &                               (Past,) &                  0.075 \\
         Sensitive &            40 &                                   37 &                                   39 &                                   40 &              2 &                 (Yes,) &                0.625 &                                 (No,) &                  0.375 \\
\bottomrule
\end{tabular}

    \caption{Annotation consistency check.}
    \label{tab:consistency}
\end{table*}

\begin{figure*}[t]
    \centering
    \includegraphics[width=0.9\textwidth]{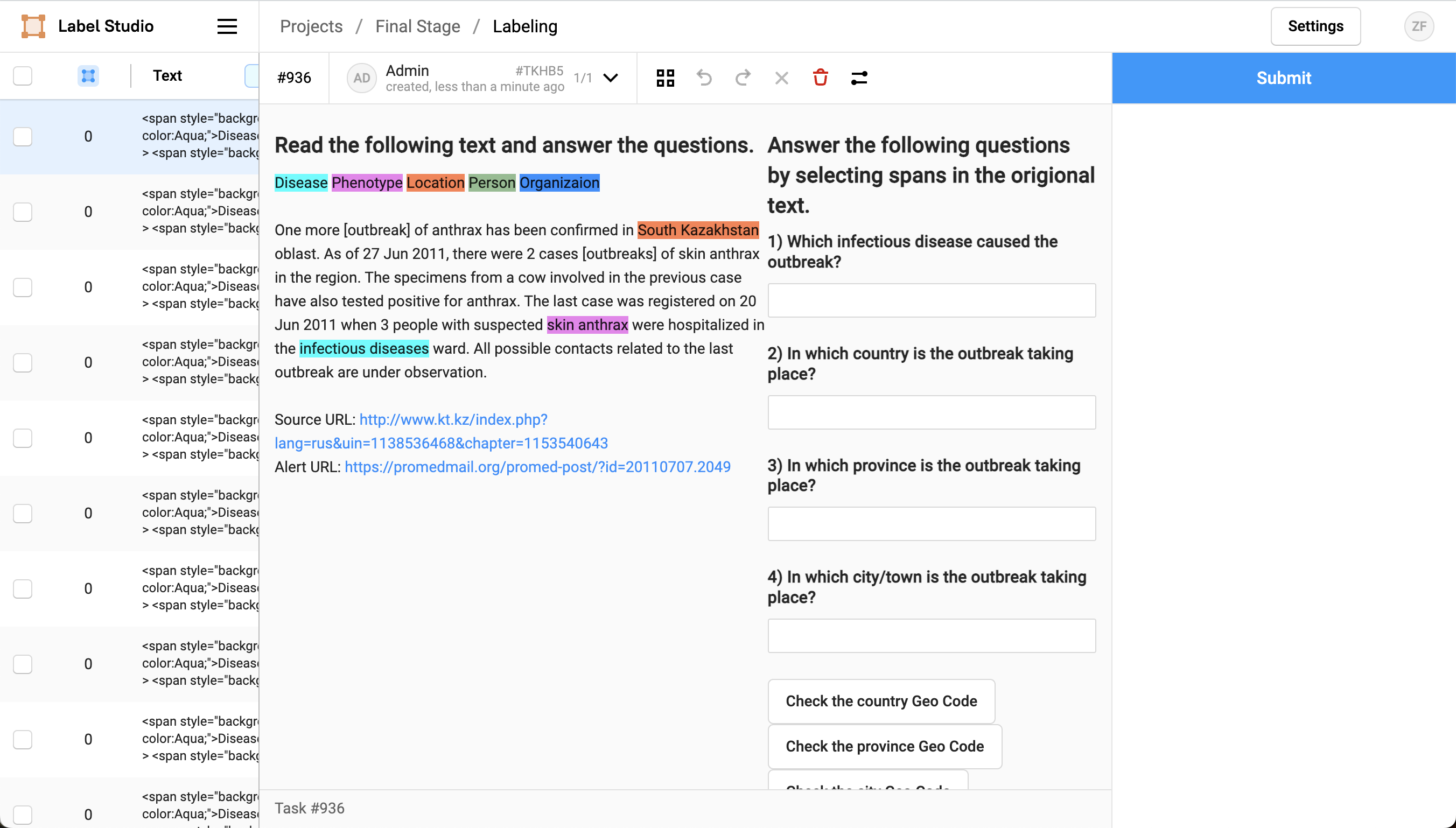}
    \caption{Annotation interface.}
    \label{fig:interface}
\end{figure*}

To guide ChatGPT in accomplishing both the QA and EE tasks, we employ specifically curated instruction sets. These facilitate accurate response generation for the QA task and pertinent event information extraction for the EE task. The instructions for the QA and EE tasks are depicted in \Cref{fig:chatinstruct} and \Cref{fig:chatinstructEE}, respectively.

\begin{figure*}

\fbox{\parbox{1\linewidth}{Read the following text and answer the following questions. If the text does not directly mention, try the best to infer from the context. If the question cannot be infered just say "Cannot Infer". The answer should be in json format.

\{New Article\}

The questions are:
1)  Which infectious disease caused the outbreak? (answer with the exact name mentioned in the given text)
2)  In which country is the outbreak taking place? (answer with the exact name mentioned in the given text, if not mentiond, please try the best to infer from the context)
3)  In which province is the outbreak taking place? (answer with the exact name mentioned in the given text, if not mentiond, please try the best to infer from the context)
4)  In which city/town is the outbreak taking place? (answer with the exact name mentioned in the given text, if not mentiond, please try the best to infer from the context)
5)  Check and fill country Geo Code (e.g. 1794299):
6)  Check and fill province Geo Code (e.g. 1794299):
7) Check and fill city Geo Code (e.g. 1815286):
8)  Which virus or bacteria caused the outbreak? Give the exact name. (answer with the exact name mentioned in the given text)
9)  What symptoms were experienced by the infected victims? (answer with the exact name mentioned in the given text)
10) Which institution reported this outbreak? (answer with the exact name mentioned in the given text)
11) What is the type of the victims? (answer with Human / Animal /Plant)
12)  How many new infected cases are reported in the specific event in the report? (please input digits like 1, 34, etc.)
13) Has the victim of the disease travelled across international borders? (answer with Yes / No / Cannot Infer)
14) Does the outbreak involve the intentful release? (answer with Yes / No / Cannot Infer)
15) Did human victims acquire the infectious disease from an animal? (answer with Yes / No / Cannot Infer)
16) Did the victim fail to respond to a drug? (answer with Yes / No / Cannot Infer)
17) Are healthcare workers included in the infected victims? (answer with Yes / No / Cannot Infer)
18) Are animal workers included in the infected victims? (answer with Yes / No / Cannot Infer)
19) Is the victim of the disease a military worker? (answer with Yes / No / Cannot Infer)
20) Did the outbreak involve a suspected contaminated blood product or vaccine? (answer with Yes / No / Cannot Infer)
21) Are the victims in a group in time and place? (answer with Yes / No / Cannot Infer)
22) Did the victim catch the disease during a hospital stay? (answer with Yes / No / Cannot Infer)
23) Is the victim of the disease a child? (answer with Yes / No / Cannot Infer)
24) Is the victim of the disease an elderly person? (answer with Yes / No / Cannot Infer)
25) Is the victim of the disease a pregnant woman? (answer with Yes / No / Cannot Infer)
26) Has the victim of the disease been in quarantine? (answer with Yes / No / Cannot Infer)
27) Did the outbreak take place during a major sporting or cultural event? (answer with Yes / No / Cannot Infer)
28) Did the outbreak take place after a natural disaster? (answer with Yes / No / Cannot Infer)
29) When did the outbreak happen? (Relative to article completion time) (answer with Now / Past)
30) Does the text contain information that can uniquely identify individual people? e.g. names, email, phone, and credit card numbers, addresses, user names. (answer with Yes / No)

Answer the questions in json format. The key shold only contain integer like "1".}}
\caption{ChatGPT instructions for QA tasks.}
\label{fig:chatinstruct}
\end{figure*}

\begin{figure*}
\fbox{\parbox{0.98\linewidth}{You are an AI expert in epidemiology. Your task is to extract information about an outbreak event from a given document. An outbreak event includes the following attributes: disease, city, province, country, city geocode, province geocode (e.g. Ohio has geocode 5165418), country geocode (e.g. United Kingdom has geocode 2635167), virus/bacteria (named as virus), symptoms (e.g. coughing), and victims (one of Human, Animal, or Plant). Note that geocodes are not in the text, so you'll need to use your background knowledge to infer them. If some attributes are not explicitly mentioned, try to infer them carefully. If the attributes cannot be found or inferred from the text, return unknown. Provide your answer in JSON format. Multiple values can be provided for the same attribute if necessary. If one attribute has multiple values, return them separated with ';'. After reading the document and receive the prefix 'Event: ', you should return: \{"disease": "replace with value\_found or unknown",\\    "city": "replace with value\_found or unknown",\\    "province": "replace with value\_found or unknown",\\    "country": "replace with value\_found or unknown",\\    "city geocode": "infer from background knowledge based on the city value found if not unknown",\\    "province geocode": "infer from background knowledge based on the province value found if not unknown",\\    "country geocode": "infer from background knowledge based on the country value found if not unknown",\\    "virus": "replace with value\_found or unknown",\\    "symptoms": "replace with value\_found or unknown",\\    "victims": "replace with value\_found or unknown"\\ \}. 

Extract the outbreak event from the following document:

\{New Article\}

Event: }}
\caption{ChatGPT instructions for EE tasks.}
\label{fig:chatinstructEE}
\end{figure*}

\begin{figure*}
\fbox{\parbox{1\linewidth}{
    \textbf{Text}

\textcolor{gray}{Las Vegas public health officials say dozens of people linked to a tuberculosis outbreak at a neonatal unit have tested positive for the disease. The Southern Nevada Health District reported on Monday  that of the 977 people tested, 59 showed indications of the disease and 2 showed signs of being contagious.\\Dr Joe Iser, chief medical officer at the health district, says the report demonstrates the importance of catching tuberculosis early. Health officials tested hundreds of babies, family members, and staff who were at Summerlin Hospital Medical Center's neonatal intensive care unit this past summer [2013], saying they wanted to take extra precautions after the death of a mother and her twin babies.\\They contacted the parents of about 140 babies who were at the unit between mid-May and mid-August [2013].}

\section{Annotation Framework}

\textbf{Questions \& Answers}

1)  Which infectious disease caused the outbreak? \textcolor{cyan}{tuberculosis} 

2)  In which country is the outbreak taking place? \textcolor{cyan}{United States} 

3)  In which province is the outbreak taking place? \textcolor{cyan}{Nevada} 

4)  In which city/town is the outbreak taking place? \textcolor{cyan}{Las Vegas} 

5)  Check and fill country Geo Code (e.g. 1794299): \textcolor{cyan}{6252001} 

6)  Check and fill province Geo Code (e.g. 1794299): \textcolor{cyan}{5509151} 

7) Check and fill city Geo Code (e.g. 1815286): \textcolor{cyan}{5506956} 

8)  Which virus or bacteria caused the outbreak? \textcolor{cyan}{Mycobacterium tuberculosis} 

9)  What symptoms were experienced by the infected victims? \textcolor{cyan}{nan} 

10) Which institution reported this outbreak? \textcolor{cyan}{Las Vegas public health} 

11) What is the type of the victims? \textcolor{cyan}{Human} 

12)  How many new infected cases are reported in the specific event in the report? (please input digits like 1, 34, etc.) \textcolor{cyan}{61.0} 

13) Has the victim of the disease travelled across international borders? \textcolor{cyan}{Cannot Infer} 

14) Does the outbreak involve the intentful release? \textcolor{cyan}{NO} 

15) Did human victims acquire the infectious disease from an animal? \textcolor{cyan}{NO} 

16) Did the victim fail to respond to a drug? \textcolor{cyan}{Cannot Infer} 

17) Are healthcare workers included in the infected victims? \textcolor{cyan}{Cannot Infer} 

18) Are animal workers included in the infected victims? \textcolor{cyan}{Cannot Infer} 

19) Is the victim of the disease a military worker? \textcolor{cyan}{Cannot Infer} 

20) Did the outbreak involve a suspected contaminated blood product or vaccine? \textcolor{cyan}{Cannot Infer} 

21) Are the victims in a group in time and place? \textcolor{cyan}{Cannot Infer} 

22) Did the victim catch the disease during a hospital stay? \textcolor{cyan}{Cannot Infer} 

23) Is the victim of the disease a child? \textcolor{cyan}{Cannot Infer} 

24) Is the victim of the disease an elderly person? \textcolor{cyan}{Cannot Infer} 

25) Is the victim of the disease a pregnant woman? \textcolor{cyan}{Cannot Infer} 

26) Has the victim of the disease been in quarantine? \textcolor{cyan}{Cannot Infer} 

27) Did the outbreak take place during a major sporting or cultural event? \textcolor{cyan}{Cannot Infer} 

28) Did the outbreak take place after a natural disaster? \textcolor{cyan}{NO} 

29) When did the outbreak happen? (Relative to article completion time) \textcolor{cyan}{Now} 

30) Does the text contain information that can uniquely identify individual people? e.g. names, email, phone, and credit card numbers, addresses, user names. \textcolor{cyan}{No}

}}
\caption{A sample with answer.}
\label{fig:sample1}
\end{figure*}

\end{document}